\documentclass[UTF8,10pt,letterpaper,fontset=fandol,scheme=plain]{ctexart}

\usepackage[left=0.8125in,top=1in,textwidth=6.875in,textheight=8.875in]{geometry}
\usepackage{amsmath,amssymb}
\usepackage{graphicx}
\usepackage{xcolor}
\usepackage{booktabs}
\usepackage{float}
\usepackage[numbers,sort&compress]{natbib}
\usepackage[format=plain,labelformat=simple,labelsep=period,font=small]{caption}
\usepackage{subcaption}
\usepackage[colorlinks=true,urlcolor=blue,linkcolor=black,citecolor=black]{hyperref}

\ctexset{
  section = {format=\large\bfseries\raggedright, name={}, number=\arabic{section},
    beforeskip=10pt plus 2pt minus 2pt, afterskip=7pt, afterindent=true},
  subsection = {format=\fontsize{11}{13.2}\selectfont\bfseries\raggedright,
    beforeskip=8pt plus 2pt minus 2pt, afterskip=5pt, afterindent=true},
  subsubsection = {format=\normalsize\bfseries\raggedright,
    beforeskip=6pt plus 2pt minus 2pt, afterskip=3pt, afterindent=true}
}
\renewenvironment{abstract}{%
  \par\centerline{\large\bfseries Abstract}%
  \vspace{12pt}\noindent\itshape\ignorespaces
}{\par\vspace{12pt}}

\newcommand{\papertitle}{RICapBench: A Fine-Grained Visual Understanding Benchmark Dataset for Vision-Language Models in Remote Sensing}
\newcommand{\paperauthors}{\large Xue Wang\textsuperscript{*,1}, Yubin Xu\textsuperscript{*,2}, Kun Tan\textsuperscript{*,\textdagger,2}\\
{\small \textsuperscript{1}the Key Laboratory of Geographic Information Science (Ministry of Education),}\\
{\small East China Normal University, Shanghai 200241, China}\\
{\small \textsuperscript{2}School of Geospatial Artificial Intelligence, East China Normal University, Shanghai, China}\\
{\small wx\_ecnu@yeah.net, 51290161018@stu.ecnu.edu.cn, tankuncu@gmail.com}\\
{\small \textsuperscript{*}Equal contribution}\\
{\small \textsuperscript{\textdagger}Corresponding author}}
\title{\Large\bfseries\papertitle}
\author{\paperauthors}
\date{}

\begin{document}
\raggedbottom
\maketitle

\begin{abstract}
Advances in vision-language models (VLMs) have improved remote sensing image analysis and supported geospatial decision-making. However, fine-grained understanding of remote sensing imagery remains challenging for VLMs.
Large variations in object scale, complex geospatial relations among instances, and limited visual detail hinder precise instance localization, visible attribute identification, and spatial relation understanding, potentially leading to descriptions that lack visual grounding.
Existing datasets and benchmarks insufficiently assess end-to-end instance localization and captioning without target-specific textual cues.
To support this evaluation, we introduce RICapBench, a dataset and benchmark that unifies instance localization and fine-grained captioning in remote sensing. RICapBench contains 245,466 instance-level annotations and 498,005 spatial relation tuples spanning general remote sensing,
UAV remote sensing, and dense small-object scenarios. Each annotation pairs an object's oriented bounding box with a natural language description of its visible attributes and spatial relations to surrounding instances and environmental elements.
To evaluate the accuracy and completeness of spatial relation captioning,
we propose a remote sensing spatial relation parser and the RelCap-F1 metric. We combine RelCap-F1 with instance localization and attribute captioning metrics into a composite metric, RICScore.
Our evaluation of 15 advanced open-source and closed-source VLMs identifies instance localization as a major bottleneck for these models.
LocateAnything achieves the highest RICScore of 31.68, but scores lower on spatial relation captioning than other task-fine-tuned open-source VLMs.
Code and dataset are available at \href{https://github.com/Xldln/RICapBench}{this URL.}
\end{abstract}
\section{Introduction}
\label{sec:introduction}

Vision-language models (VLMs) have become an important research direction in deep learning and computer vision in recent years~\cite{cho2025perceptionlm,xing2025conical}.
Vision and language underpin human perception of the external world and the expression of knowledge, respectively. Integrating them enables models to understand visual content and interact through natural language, supporting tasks such as visual question answering, image captioning, and object localization~\cite{liu2023visual,liu2024improved}.

VLMs have attracted considerable attention in remote sensing, with growing applications in object detection and segmentation, change detection, and damage assessment for urban planning and geospatial analysis~\cite{kuckreja2024geochat,pang2025vhm,soni2025earthdial,li2025segearthov,shabbir2025geopixel,wang2024earthvqa}.
Despite their potential to reduce manual effort and improve analytical efficiency, fine-grained understanding remains challenging. Large scale variations and densely distributed small objects with limited discernible detail complicate instance identification, counting, and precise localization~\cite{zhang2025geovis}.
Beyond instance localization and description, models must distinguish adjacent objects and determine their positions relative to surrounding instances and environmental elements using scene context. Image-level captioning alone does not ensure accurate understanding of these spatial relations.
Models may also hallucinate objects, attributes, and spatial relations that contradict image content or lack visual grounding, potentially undermining automated interpretation reliability~\cite{an2025choice,pang2025vhm}. Together, these challenges demand stronger fine-grained image understanding.

Remote sensing benchmarks such as DIOR-RSVG~\cite{zhan2023rsvg}, VRSBench~\cite{li2024vrsbench}, EarthVQA~\cite{wang2024earthvqa}, and GEOBench-VLM~\cite{danish2025geobenchvlm} cover image captioning, visual question answering, and referring expression grounding.
These tasks assess instance-level understanding through language--object correspondence and spatial relation reasoning.
However, referring expression grounding typically relies on target descriptions whose references, attributes, or relations guide attention to specific objects~\cite{lai2024lisa}. Performance under such guidance therefore does not fully reflect models' autonomous fine-grained understanding without target-specific cues.
Clearly describing object appearance and location requires substantial manual effort~\cite{wang2024omniq}. Constructing queries may require users to specify targets or inspect imagery beforehand, adding interaction overhead that may delay UAV-based disaster inspection and remote sensing emergency monitoring.
Meanwhile, image-level captioning generally does not require descriptions to be matched to specific instance locations, limiting its assessment of instance attributes and spatial relations.

In general vision, previous studies have combined object localization with fine-grained language descriptions. For example, Dense Relational Captioning~\cite{kim2019drc} focuses on describing relations between objects, whereas LVISCap~\cite{fiastre2026captionformer} provides fine-grained object-level descriptions.
RA-4M~\cite{neau2026relateanything} further provides annotations of inter-object relations expressed as free-text predicates. These studies inform the joint assessment of object localization, object captioning, and inter-object relation understanding.
In remote sensing, FIT-RS~\cite{luo2024skysensegpt} supports region captioning conditioned on target regions and whole-image scene graph generation. The latter uses templated sentences with explicit object categories, locations, and relation predicates.
FineGrip~\cite{zhao2024finegrip} combines segmentation with image-level captioning, without directly evaluating instance--caption correspondence or separately assessing the accuracy and completeness of spatial relation captioning.
We investigate whether VLMs can localize remote sensing instances and describe their visible attributes and spatial relations to surrounding instances and environmental elements without target-specific text queries.

To systematically evaluate fine-grained understanding under this setting, we introduce RICapBench, which unifies instance localization, attribute captioning, and spatial relation captioning within a single evaluation framework. This framework supports the analysis of capability differences and bottlenecks in general-purpose and remote sensing VLMs, as well as the exploration of potential improvements.
Using RICapBench, we systematically evaluate 15 advanced general-purpose and remote sensing VLMs on fine-grained understanding of remote sensing imagery. The task examines whether models can localize target instances based on specific visual evidence and describe their attributes and spatial relations to surrounding geographic features or environmental elements.
The results show that existing general-purpose and remote sensing VLMs continue to face challenges in fine-grained understanding of remote sensing instances and their spatial context.

Our main contributions are summarized as follows:
\begin{itemize}
    \item We introduce RICapBench with 245,466 instance annotations and 498,005 spatial relation tuples across general remote sensing, UAV remote sensing, and dense small-object scenarios. Annotations pair oriented bounding boxes with attribute and spatial relation descriptions, supporting fine-tuning and evaluation without target-specific text queries.
    \item We develop a remote sensing spatial relation parser that extracts structured triples from captions and propose RelCap-F1 to measure relation captioning accuracy and completeness. Our composite metric, RICScore, combines DetA, CapA, and RelCap-F1 through their geometric mean to jointly assess localization, attribute captioning, and spatial relation captioning.
    \item We evaluate 15 VLMs, covering general-purpose open-source, closed-source, and remote sensing models, and analyze their capability differences and bottlenecks across the three evaluation dimensions.
\end{itemize}
\section{Related Work}
\label{sec:overview}

Table~\ref{tab:benchmark-comparison} compares RICapBench with related datasets and benchmarks for fine-grained visual understanding in general vision and remote sensing.

\begin{table}[!htb]
    \caption{
        Comparison of RICapBench with related vision--language datasets and benchmarks in general vision and remote sensing.
        For FIT-RS, the sample count refers to instruction samples, and only selected tasks are listed.
        Detailed explanations (HBB: Horizontal Bounding Box, OBB: Oriented Bounding Box, Seg: segmentation, Loc.: localization, A: Automatic, M: Manual,
        IS: instance segmentation, VG: visual grounding, VQA: visual question answering, SGG: scene graph generation, IC: image captioning, RC: region captioning, PS: panoptic segmentation, RIS: referring image segmentation;
        OD: object detection, RP: relation prediction, RR: relation reasoning, RelCap: relational captioning)}
    \label{tab:benchmark-comparison}
    \centering
    \small
    \setlength{\tabcolsep}{3pt}
    \renewcommand{\arraystretch}{1.15}
    \begin{tabular*}{\textwidth}{@{\extracolsep{\fill}}ccccccc@{}}
        \toprule
        Dataset / Benchmark & Year & Samples & Localization & Annotation Type & Granularity & Tasks \\
        \midrule
        \multicolumn{7}{@{}l@{}}{\textcolor{black!65}{\textit{General vision}}} \\
        \begin{tabular}[c]{@{}c@{}}Dense Relational\\Captioning~\cite{kim2019drc}\end{tabular} & 2019 & --- & HBB & A & Instance & Loc. + RelCap \\
        LVISCap~\cite{fiastre2026captionformer} & 2025 & 1.5M & HBB, Seg & A & Instance & IS + Cap. \\
        RA-4M~\cite{neau2026relateanything} & 2026 & 4.3M & --- & A & Instance & RP \\
        \midrule
        \multicolumn{7}{@{}l@{}}{\textcolor{black!65}{\textit{Remote sensing}}} \\
        DIOR-RSVG~\cite{zhan2023rsvg} & 2022 & 38.3K & HBB & A+M & Instance & VG \\
        EarthVQA~\cite{wang2024earthvqa} & 2023 & 208.6K & --- & A+M & Image / Instance & VQA \\
        FineGrip~\cite{zhao2024finegrip} & 2024 & 13.2K & Seg & A+M & Image & PS + IC \\
        FIT-RS~\cite{luo2024skysensegpt} & 2024 & 1.8M & OBB & A+M & \begin{tabular}[c]{@{}c@{}}Image / Region\\Instance\end{tabular} & IC, RC, VQA, OD, RR, SGG, \ldots \\
        RRSIS-D~\cite{liu2024rmsin} & 2023 & 17.4K & Seg & A+M & Instance & RIS \\
        VRSBench~\cite{li2024vrsbench} & 2024 & 205.3K & HBB / OBB & A+M & Image / Instance & IC, VG, VQA \\
        AerialVG~\cite{liu2025aerialvg} & 2025 & $\sim$50.0K & HBB & M & Instance & VG \\
        RSICap / RSIEval~\cite{hu2025rsgpt} & 2023 & 3.6K & --- & M & Image / Instance & IC, VQA \\
        \midrule
        RICapBench (Ours) & 2026 & 245.5K & OBB & A+M & Instance & Loc. + RelCap \\
        \bottomrule
    \end{tabular*}
\end{table}

\subsection{General VLM Benchmarks}
Several datasets and benchmarks focus on fine-grained visual understanding, providing a foundation for evaluating VLMs' understanding of objects and their context.
Fiastre et al.~\cite{fiastre2026captionformer} extend LVIS and LV-VIS to create LVISCap and LV-VISCap, respectively, providing fine-grained object-level descriptions for images and videos.
These datasets support combining object detection, segmentation, and video object tracking with object captioning. Their descriptions cover object appearance, actions, and interactions with the surrounding environment or other objects.
Benchmarks such as SEED-Bench~\cite{li2023seedbench} and MMStar~\cite{chen2024mmstar} evaluate fine-grained perception through visual question answering. They cover complex perceptual tasks involving object attribute identification, object localization, relation reasoning, and interactions between instances.
These benchmarks focus on general visual scenes rather than spatial relation captioning between geographic feature instances in remote sensing imagery.

\subsection{Domain-Geospatial Benchmarks}
\begingroup
\emergencystretch=2em
VRSBench~\cite{li2024vrsbench} covers image captioning, visual question answering, and visual grounding. RSIEval~\cite{hu2025rsgpt}, introduced in RSGPT, focuses on image captioning and visual question answering. LHRS-Bench~\cite{muhtar2025lhrsbot} uses multiple-choice questions to evaluate object recognition, spatial perception, and reasoning.
EarthVQA~\cite{wang2024earthvqa} focuses on relation reasoning between geographic objects and comprehensive scene analysis, primarily for urban and rural planning and governance. GeoChat-Instruct~\cite{kuckreja2024geochat} supports image-level and region-level remote sensing dialogue, emphasizing the correspondence between natural language descriptions and object locations.
Visual question answering and visual grounding typically guide attention to specific objects and spatial relations through category, attribute, or location cues in questions or referring expressions.

\par\endgroup
\section{RICapBench Dataset}
\label{sec:benchmark}
RICapBench contains 26,854 remote sensing images, each measuring $800 \times 800$ pixels, with 245,466 instance-level annotations and 498,005 spatial relation tuples.
The images are sourced from DIOR-R~\cite{cheng2022aopg}, CODrone~\cite{ye2025codrone}, and FAIR1Mv2.0~\cite{sun2022fair1m}, representing general remote sensing, UAV remote sensing, and dense small-object scenarios, respectively. The oriented bounding boxes retain the original annotations from each source dataset.
Building on these annotations, we generate fine-grained natural language descriptions of target instances through a workflow combining semi-automatic annotation with manual verification. Each annotation contains the object's oriented bounding box, category, and corresponding description.
RICapBench integrates remote sensing images from diverse scenarios to support the evaluation of VLMs' understanding of instances and their spatial relations across imaging perspectives, object scales, and distribution densities.

\subsection{Annotation Schema}
\begin{figure}[!htb]
    \centering
    \includegraphics[width=0.82\textwidth]{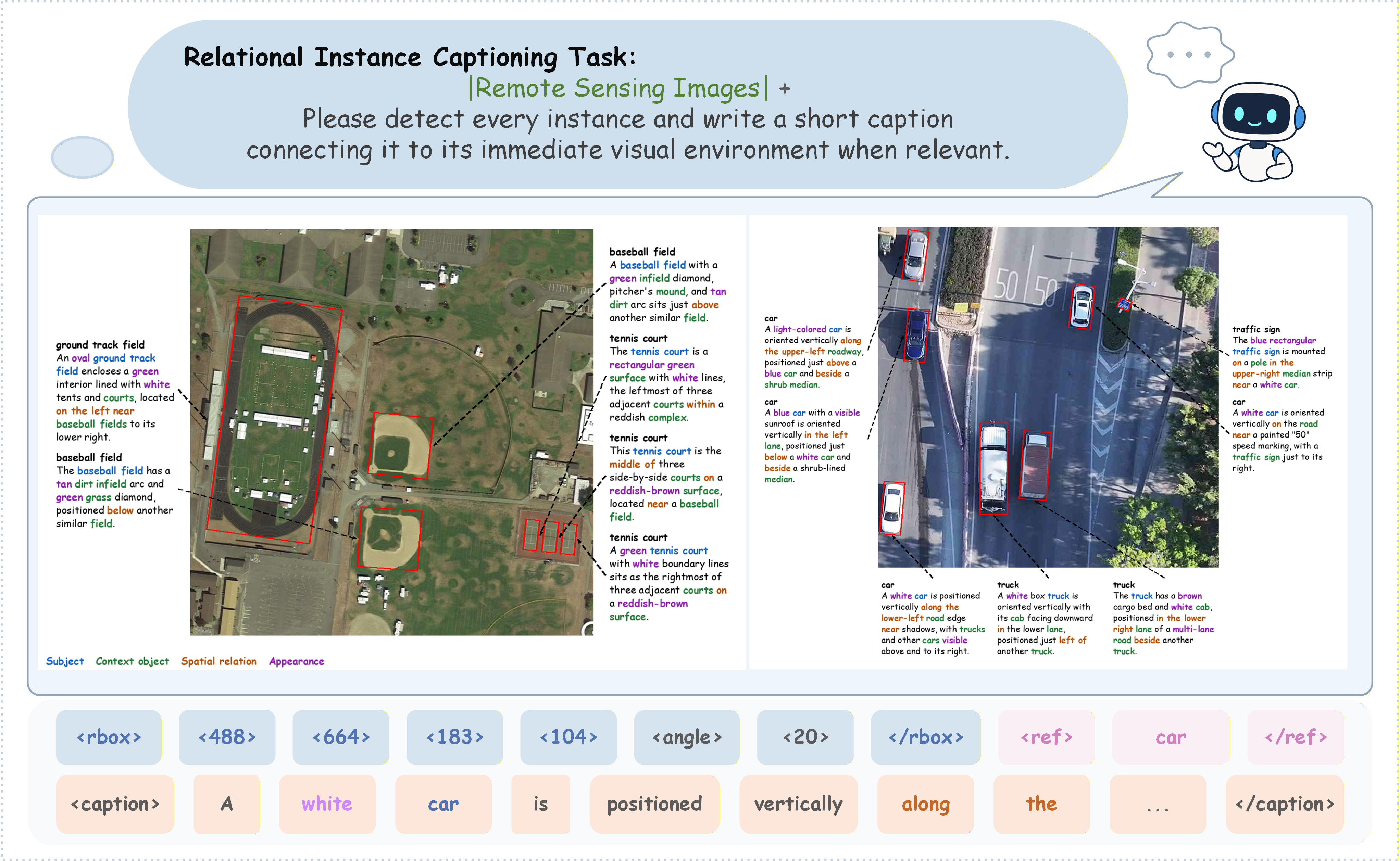}
    \caption{Overview of the remote sensing instance spatial relationship captioning task and the annotation format and examples of RICapBench.}
    \label{fig:annotation-schema}
\end{figure}

RICapBench normalizes the coordinates and dimensions of oriented bounding boxes and represents them uniformly as $(c_x,c_y,w,h,\theta)$. Here, $(c_x,c_y)$ denotes the box center coordinates, $w$ and $h$ denote its width and height, respectively, and $\theta$ denotes the rotation angle.
Annotation examples are shown in the lower part of Figure~\ref{fig:annotation-schema}. Each instance-level annotation comprises three fields: \texttt{<rbox>}, \texttt{<ref>}, and \texttt{<caption>}, encoding the oriented bounding box, instance category, and corresponding natural language description, respectively.

Instance descriptions cover the object's appearance attributes and visible state, its location within the image, and its spatial relations to surrounding environmental elements and other instances.
Appearance attributes and visible state capture the object's own visual characteristics, such as color, size, and discernible state. The image location specifies where the object appears in the current image. Spatial relations describe its position and spatial arrangement relative to environmental elements or other instances.
Spatial relation descriptions take the target instance as the subject. Reference objects include other annotated instances in the image and discernible environmental elements without separate instance annotations. All descriptions are grounded in discernible visual information in the image. When no clearly identifiable reference objects surround the target, descriptions focus on its own visible features and location within the image.

\subsection{Construction and Quality Control}

Research and evaluation on fine-grained vision-language tasks rely on high-quality annotated data. Annotation assisted by large language models and VLMs offers a scalable approach to constructing such datasets.
For example, LLaVA-Grounding~\cite{zhang2024llavagrounding} and GLaMM~\cite{rasheed2024glamm} use large models to assist in constructing annotated data that combine visual grounding with language descriptions. In remote sensing, VRSBench~\cite{li2024vrsbench} and GeoChat~\cite{kuckreja2024geochat} also employ model-assisted annotation workflows to construct vision-language datasets. Following these studies, we construct RICapBench through a workflow that combines model-assisted generation with manual verification.
To ensure that instance descriptions cover both the target's own visible features and its spatial relations to surrounding instances and environmental elements, we guide caption generation using global and local visual prompts together with textual constraints. We then manually verify the visual grounding of the descriptions and the accuracy of the spatial relations.

\begin{figure}[!htb]
    \centering
    \includegraphics[width=0.88\textwidth]{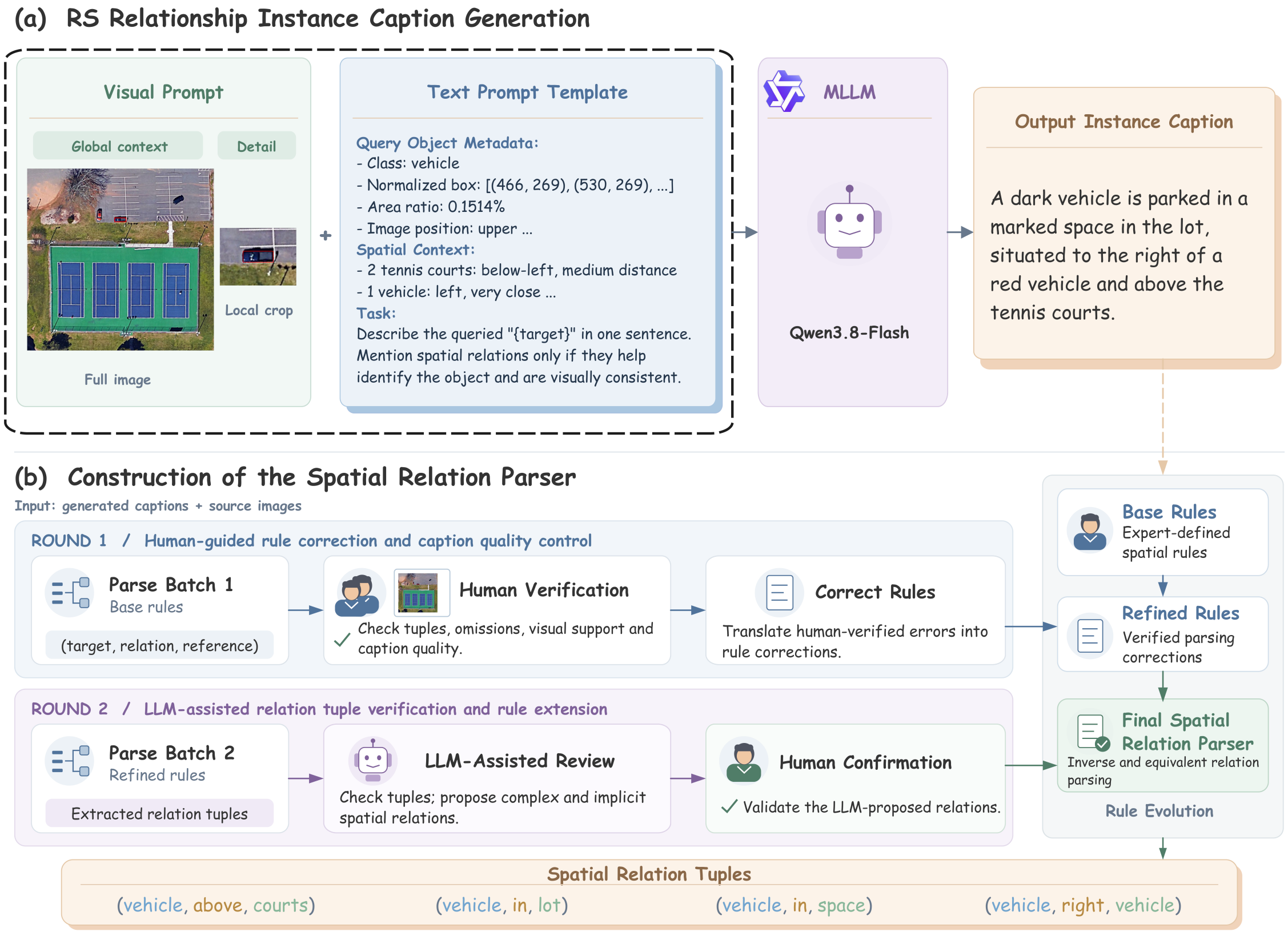}
    \caption{Pipelines for (a) generating remote sensing instance captions with spatial relationships and (b) constructing the spatial relation parser.}
    \label{fig:construction-pipeline}
\end{figure}

\textbf{Prompt Template.}
We combine visual prompts with textual constraints to generate instance-level captions for RICapBench using Qwen3.8-Flash~\cite{alibaba2026qwen38flash}, a general-purpose large multimodal model. As shown in Figure~\ref{fig:construction-pipeline}, we highlight the target in the global image using its oriented bounding box from the source dataset. We also provide a magnified local view of the target region.
The global image preserves the spatial context linking the target to surrounding instances and environmental elements. The magnified local view helps identify the target's appearance details and visible state. The text prompts comprise a global system prompt and an instance-level prompt template, specifying the caption content, output format, and visual grounding requirements.
These prompts guide Qwen3.8-Flash to integrate global and local information, reduce inferences unsupported by visual evidence, and improve caption accuracy and consistency.

\textbf{Spatial Relationship Parser.}
The image captioning metric SPICE~\cite{anderson2016spice} uses dependency parsing and rule-based semantic parsing to convert candidate and reference captions into scene graphs containing objects, attributes, and inter-object relations. However, its general-purpose relation parsing outputs may not meet the representation requirements for evaluating spatial relations of remote sensing instances.
For example, applying SPICE to the output instance caption in Figure~\ref{fig:construction-pipeline}(a) yields four subject--predicate--object relation triples: (lot, situate above, court), (lot, situate to, right), (right, of, vehicle), and (space, in, lot).
The triples (lot, situate to, right) and (right, of, vehicle) treat the directional term \emph{right} as a relation endpoint, splitting a complete directional expression into two relations. Consequently, neither triple directly represents the target's position relative to the reference geographic feature.
Relation tuples in general-purpose scene graphs therefore require further processing to preserve complete spatial expressions and ensure the validity of reference objects. To meet the requirements for parsing and evaluating spatial relations of remote sensing instances, we develop a spatial relation parser tailored to these instances.

As shown in Figure~\ref{fig:construction-pipeline}(b), domain experts first establish a set of initial parsing rules. These include rules for relation triple extraction, spatial phrase matching, and a whitelist of reference objects. The rules are then refined through two rounds of validation.
In the first round, the initial rules are applied to a batch of samples. Human reviewers consult the original images to verify the accuracy of spatial relation triples and add relations missed during parsing. They also assess caption quality and visual grounding. Confirmed parsing issues inform revisions and additions to the rules.
In the second round, the revised rules are applied to another batch of samples. Qwen3.8-Max~\cite{qwen2026qwen38max} assists in checking relation triples and suggests refinements for parsing inverse relations, symmetric relations, and other complex sentence structures.
All candidate relations that Qwen3.8-Max proposes to remove or add are reviewed and confirmed by human reviewers. Verified parsing issues and suggestions are then incorporated into rule revisions to produce the final spatial relation parser.

\subsection{Dataset Statistics}

\begin{figure}[!htb]
    \centering
    \captionsetup[subfigure]{font=small,justification=centering,singlelinecheck=false,skip=5pt}
    \begin{subfigure}[t]{0.34\textwidth}
        \centering
        \vspace{0pt}
        \parbox[c][1.85in][c]{\linewidth}{\centering
            \includegraphics[trim=6bp 1bp 26bp 1bp,clip,width=\linewidth,height=1.85in,keepaspectratio]{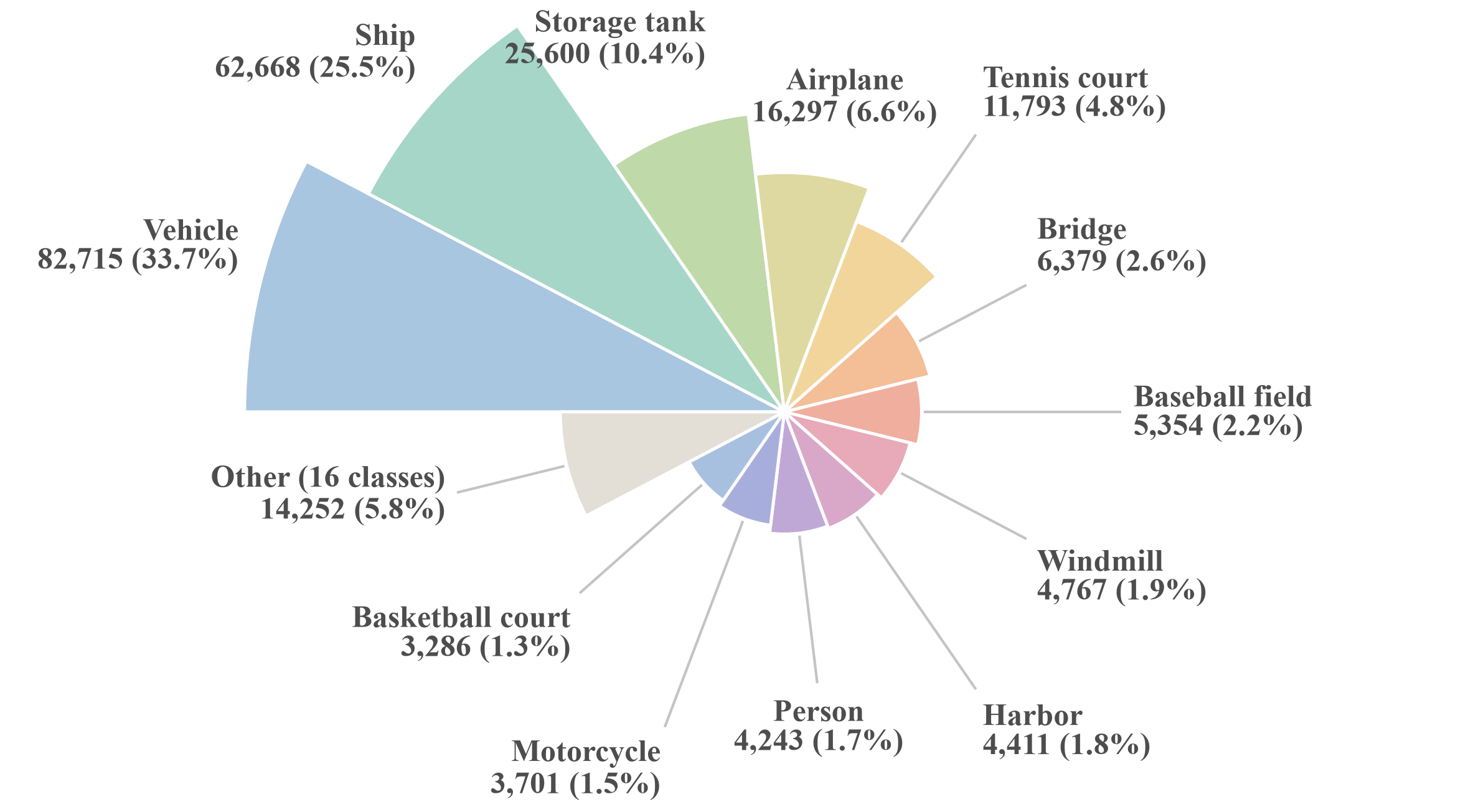}}
        \caption{Object categories}
        \label{fig:category-distribution}
    \end{subfigure}\hfill
    \begin{subfigure}[t]{0.27\textwidth}
        \centering
        \vspace{0pt}
        \parbox[c][1.85in][c]{\linewidth}{\centering
            \includegraphics[trim=35bp 1bp 45bp 7bp,clip,width=\linewidth,height=1.85in,keepaspectratio]{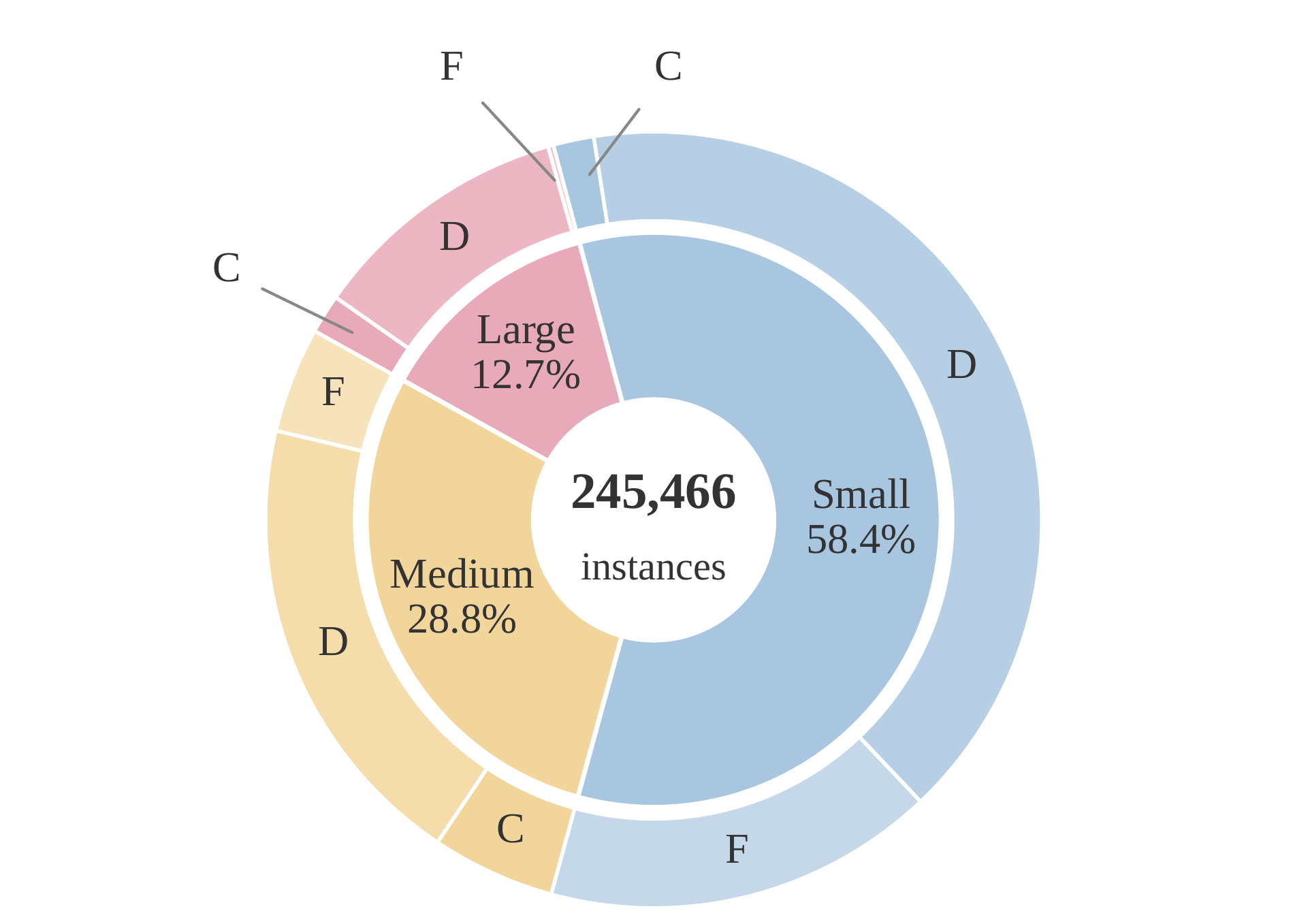}}
        \caption{Object sizes}
        \label{fig:size-distribution}
    \end{subfigure}\hfill
    \begin{subfigure}[t]{0.35\textwidth}
        \centering
        \vspace{0pt}
        \parbox[c][1.85in][c]{\linewidth}{\centering
            \includegraphics[width=\linewidth,height=1.85in,keepaspectratio]{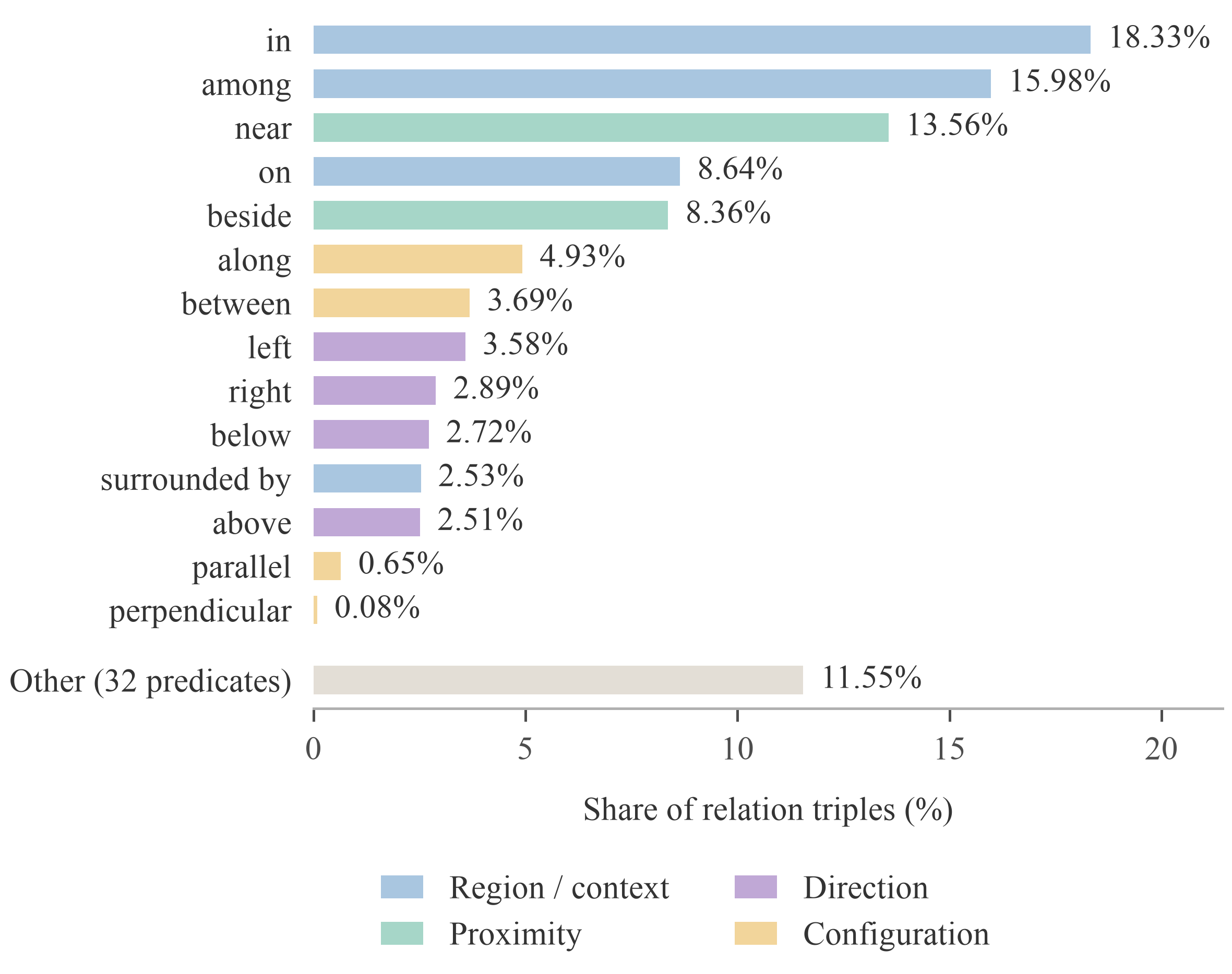}}
        \caption{Spatial predicates}
        \label{fig:spatial-predicate-coverage}
    \end{subfigure}
    \caption{Overview of RICapBench dataset statistics.
    (a) Instance counts and proportions for the 12 most frequent object categories and the remaining 16 categories grouped as Other.
    (b) The inner ring shows the proportions of small, medium, and large instances, while the outer ring shows the data source composition within each size group. C, D, and F denote CODrone, DIOR-R, and FAIR1Mv2.0, respectively.
    (c) Proportions of spatial predicates across all relation triplets, with the remaining predicates grouped as Other.}
    \label{fig:dataset-statistics}
\end{figure}

We analyze the composition of RICapBench in terms of object categories and scales, spatial relation types, and reference-object expressions. The main statistics are shown in Figure~\ref{fig:dataset-statistics}.
RICapBench retains the category annotations from its source datasets, which originally comprise 51 categories. We merge synonymous or closely related labels into 28 categories. This mapping is applied to references to surrounding annotated instances in captions to reduce naming differences and semantic inconsistencies across data sources.
Figure~\ref{fig:category-distribution} shows the instance counts for the main categories.
As shown in Figure~\ref{fig:size-distribution}, object sizes are categorized according to the MS COCO~\cite{lin2014mscoco} criteria. Small objects dominate the dataset, accounting for approximately 58.4\% of all instances, while the outer ring further shows the data source composition within each size group.
Reference-object expressions are shown in Figure~\ref{fig:spatial-reference-expressions} in the supplementary material.
Figure~\ref{fig:spatial-predicate-coverage} shows the distribution of predicates in the relation triples extracted by the spatial relation parser. The most frequent predicates are \emph{in}, \emph{among}, and \emph{near}, and the relations span direction, proximity, and spatial layout.
\section{VLMs Benchmarking}
\label{sec:evaluation}

\subsection{Model Selection}
\textbf{Open-Source Models.}
The general-purpose models include Qwen3-VL~\cite{bai2025qwen3vl}, InternVL3.5~\cite{wang2025internvl35}, Ovis2.5~\cite{lu2025ovis25}, Gemma-3~\cite{gemmateam2025gemma3}, LLaVA-OneVision~\cite{an2025llavaonevision15,an2026llavaonevision2}, and Qwen3.8-Max~\cite{qwen2026qwen38max}.
We also include LocateAnything~\cite{wang2026locateanything}, which focuses on object localization, to analyze differences between its instance localization and captioning capabilities. The remote sensing models include GeoChat~\cite{kuckreja2024geochat}, GeoPixel~\cite{shabbir2025geopixel}, and EarthDial~\cite{soni2025earthdial}.
Qwen3.8-Max is evaluated through API calls, whereas all other models in this group are evaluated after fine-tuning on the RICapBench training set using LoRA~\cite{hu2022lora}.

\textbf{Proprietary Models.}
In addition to open-source models, we evaluate Gemini-3.8-Flash~\cite{deepmind2026gemini38flash} and GPT-6-Luna~\cite{openai2026gpt6solandluna}, two closed-source general-purpose large multimodal models.

\subsection{Evaluation Metrics}

\label{sec:metrics}

Following prior work on DVOC, we build a composite metric for remote sensing instance spatial relation captioning based on the CHOTA evaluation framework proposed by Zhou et al.~\cite{zhou2025densevoc}.
We retain its detection accuracy metric, DetA, and modify the calculation of captioning accuracy, CapA. Specifically, we replace SPICE All in CapA with SPICE-OA, which includes only the object and attribute components.
We also replace CIDEr with CIDEr-D~\cite{vedantam2015cider} and divide its score by 10 before including it in the calculation. Together with METEOR~\cite{banerjee2005meteor}, these metrics form the modified CapA, calculated as follows:

\begin{equation}
\mathcal{M}_{\mathrm{cap}}
= \{(p,g)\in\mathcal{M}: |S_g|>0\}.
\end{equation}

\begin{equation}
\begin{aligned}
q_{pg}=\frac{1}{3}\Bigl[&\mathrm{METEOR}(\hat{c}_p,S_g)\\
&+\frac{\mathrm{CIDEr\text{-}D}(\hat{c}_p,S_g)}{10}\\
&+\mathrm{SPICE\text{-}OA}(\hat{c}_p,S_g)\Bigr].
\end{aligned}
\end{equation}

\begin{equation}
\mathrm{CapA}
=\frac{1}{|\mathcal{M}_{\mathrm{cap}}|}
\sum_{(p,g)\in\mathcal{M}_{\mathrm{cap}}}q_{pg}.
\end{equation}

Here, $\mathcal{M}$ is the set of matched predicted and ground-truth instances, and $\hat{c}_p$ is the caption for predicted instance $p$. The set $S_g$ contains the reference captions for ground-truth instance $g$. The set $\mathcal{M}_{\mathrm{cap}}$ retains only matched pairs with nonempty reference caption sets, and $q_{pg}$ is the caption quality score for each matched pair.

We further propose RelCap-F1 to measure the accuracy and completeness of spatial relation captioning for target instances, as shown in Equation~\eqref{eq:relcap-f1}.
We define the Relational Instance Caption Score (RICScore). It jointly measures performance in instance localization, attribute captioning, and spatial relation captioning, as shown in Equation~\eqref{eq:ricscore}.

\begin{equation}
\label{eq:relcap-f1}
\mathrm{RelCap\text{-}F1}
=\frac{2TP_{\mathrm{rel}}}
{2TP_{\mathrm{rel}}+FP_{\mathrm{rel}}+FN_{\mathrm{rel}}}.
\end{equation}

\begin{equation}
\label{eq:ricscore}
\mathrm{RICScore}
=\sqrt[3]{
\mathrm{DetA}\cdot\mathrm{CapA}\cdot\mathrm{RelCap\text{-}F1}
}.
\end{equation}

Here, $TP_{\mathrm{rel}}$, $FP_{\mathrm{rel}}$, and $FN_{\mathrm{rel}}$ denote the numbers of correctly matched, incorrectly predicted, and missed spatial relation triples, respectively.

\subsection{Benchmarking Overview}
All evaluation images come from the original validation splits of the three source datasets. We manually inspect a subset of candidate samples and exclude those with low-quality annotations, substantial image overlap, or issues such as blur and ghosting.
The final evaluation set contains 1,930 images, 16,605 instance-level annotations, and 35,774 spatial relation triples. Instance captions average approximately 21 words, and the set covers general remote sensing, UAV remote sensing, and dense-object scenarios. The training and evaluation sets are disjoint at the image level.
\section{Evaluations and Analysis}

\label{sec:results}
\subsection{Overall Results}

\begin{table}[!htb]
    \caption{Overall results on RICapBench. Adapt.: FT = fine-tuned on the RICapBench training set. RICScore is reported on a 0--100 scale,
    while all other metrics are expressed as percentages. RelCap-F1 denotes RelCap-$\mathrm{F1}_{\mathrm{matched}}$.
    }
    \label{tab:overall-results}
    \centering
    \small
    \setlength{\tabcolsep}{3pt}
    \renewcommand{\arraystretch}{1.2}
    \resizebox{\textwidth}{!}{%
    \begin{tabular}{@{}ccc|ccccccc|c@{}}
        \toprule
        Model & Params & Adapt. & METEOR & CIDEr-D/10 & SPICE-OA & RelCap-$\mathrm{F1}_{\mathrm{all}}$ & DetA & CapA & RelCap-F1 & RICScore \\
        \midrule
        Qwen3-VL-Instruct~\cite{bai2025qwen3vl}& 2B & FT & 30.22 & 20.58 & 47.15 & 8.38 & 19.74 & 32.65 & 27.51 & 26.08 \\
        InternVL3.5~\cite{wang2025internvl35} & 2B & FT &29.52 & 19.73 & 46.09 & 6.40 & 15.42 & 31.78 & 27.19 & 23.71 \\
        Ovis2.5~\cite{lu2025ovis25} & 2B & FT & 29.37 & 19.60 & 46.08 & 5.38 & 12.41 & 31.69 & 27.84 & 22.20 \\
        Gemma-3-it~\cite{gemmateam2025gemma3}& 4B & FT & 29.39 & 19.91 & 46.09 & 2.61 & 6.02 & 31.79 & 28.97 & 17.70 \\
        \shortstack[c]{LLaVA-OneVision-1.5\\-Instruct~\cite{an2025llavaonevision15}}&4B & FT & 28.93 & 19.04 & 45.25 & 2.39 & 5.50 & 31.07 & 27.62 & 16.77 \\
        Qwen3-VL-Instruct~\cite{bai2025qwen3vl}&8B   & FT & 30.17 & 20.15 & 46.70 & 10.00 & 24.63 & 32.34 & 27.15 & 27.86 \\
        InternVL3.5~\cite{wang2025internvl35} &8B   & FT & 29.68 & 19.65 & 45.89 & 7.91 & 18.57 & 31.74 & 27.45 & 25.30 \\
        \shortstack[c]{LLaVA-OneVision-2\\-Instruct~\cite{an2026llavaonevision2}}&8B & FT & 30.37 & 21.35 & 48.06 & 4.44 & 9.16 & 33.26 & 30.76 & 21.08 \\
        LocateAnything~\cite{wang2026locateanything}&3B & FT & 28.96 & 18.03 & 44.58 & 14.03 & 44.50 & 30.52 & 23.40 & 31.68 \\
        \midrule
        GeoChat~\cite{kuckreja2024geochat}& 7B & FT & 26.61 & 16.87 & 41.59 & 1.18 & 3.27 & 28.35 & 25.56 & 13.34 \\
        GeoPixel~\cite{shabbir2025geopixel}& 7B & FT & 28.53 & 19.14 & 45.00 & 2.43 & 5.91 & 30.89 & 28.64 & 17.36 \\
        EarthDial~\cite{soni2025earthdial}& 4B & FT & 28.10 & 18.18 & 44.12 & 1.85 & 4.78 & 30.13 & 26.30 & 15.59 \\
        \midrule
        Qwen3.8-Max~\cite{qwen2026qwen38max}&2.4T & API & 21.00 & 7.61 & 27.99 & 1.81 & 21.59 & 18.87 & 4.97 & 12.65 \\
        GPT-6-Luna~\cite{openai2026gpt6solandluna}& - & API & 18.05 & 5.78 & 26.00 & 0.72 & 6.58 & 16.61 & 5.98 & 8.68 \\
        Gemini-3.8-Flash~\cite{deepmind2026gemini38flash}& - & API & 14.60 & 3.31 & 25.86 & 1.41 & 17.47 & 14.59 & 5.37 & 11.10 \\
        \bottomrule
    \end{tabular}%
    }
\end{table}

Table~\ref{tab:overall-results} presents the overall performance of the evaluated VLMs on RICapBench.
LocateAnything achieved the highest composite score, with a RICScore of 31.68. The remote sensing models GeoChat, GeoPixel, and EarthDial all scored lower than most task-fine-tuned general-purpose open-source VLMs.
LocateAnything's main advantage lay in instance localization, with a DetA of 44.50\%, well above the second-ranked Qwen3-VL-8B-Instruct (24.63\%).
Even the remote sensing models achieved DetA scores below 6\%, indicating that adaptation to remote sensing does not necessarily confer an instance localization advantage on this task. These results indicate that localizing specific instances remains a major bottleneck for most VLMs in fine-grained understanding.
The task-fine-tuned open-source VLMs performed similarly in instance attribute captioning, with CapA scores ranging from 28.35\% to 33.26\%.
LLaVA-OneVision-2-8B achieved the highest RelCap-F1 of 30.76\%, but its DetA was only 9.16\%. LocateAnything, which performed best in localization, achieved a RelCap-F1 of 23.40\%, the lowest among the task-fine-tuned open-source models.
This indicates that strong instance localization does not necessarily coincide with accurate spatial relation understanding.

GPT-6-Luna, Gemini-3.8-Flash, and Qwen3.8-Max achieved RICScore values of 8.68, 11.10, and 12.65, respectively. Under the current evaluation setting, all three scored below task-fine-tuned open-source VLMs with fewer parameters.

\subsection{Analysis}
\label{sec:analysis-results}

\begin{figure}[!htb]
    \centering
    \includegraphics[width=0.72\textwidth]{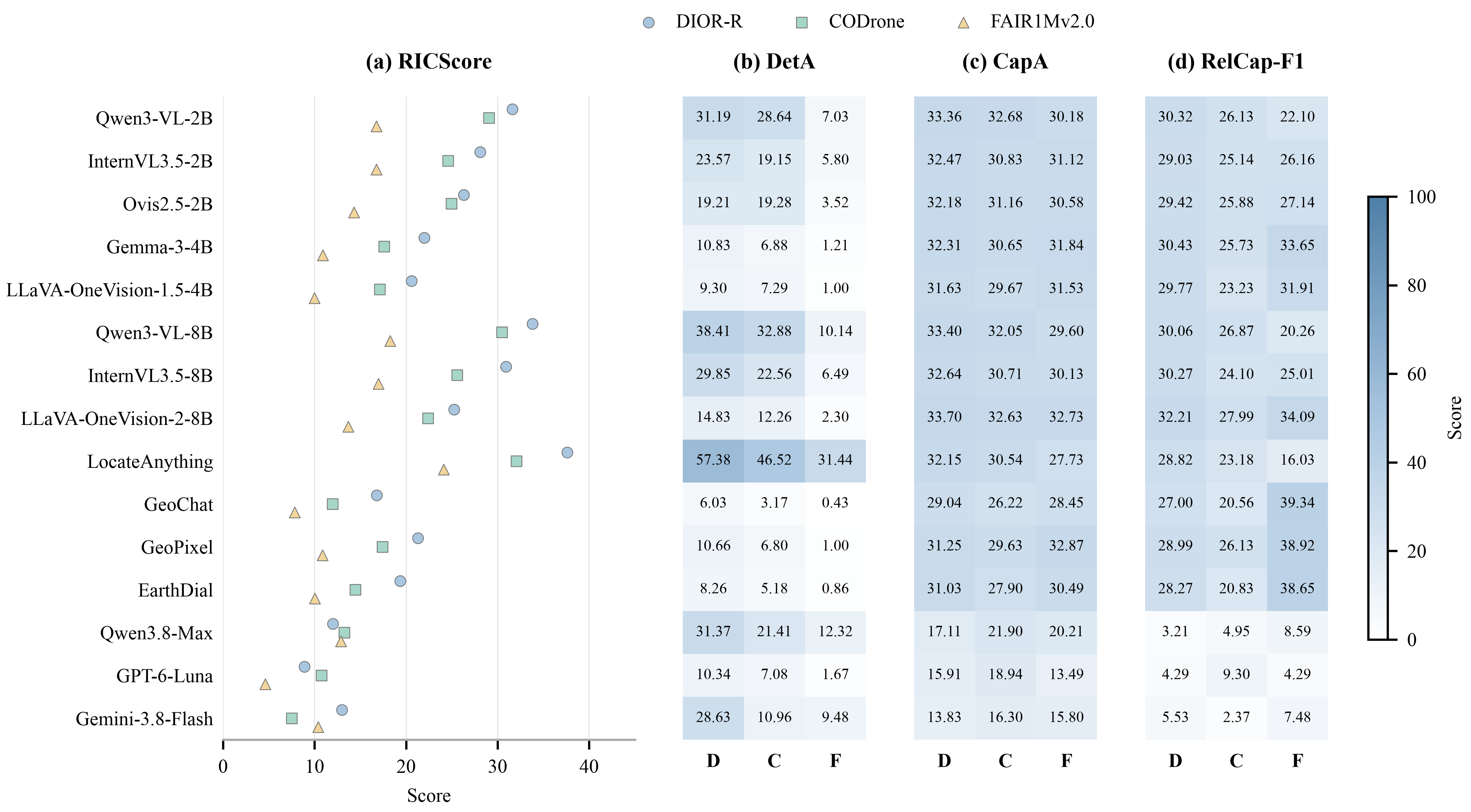}
    \caption{
    Performance of different models across three data sources.
    D, C, and F denote DIOR-R, CODrone, and FAIR1Mv2.0, respectively.
    The heatmaps share a common color scale, with darker colors indicating higher scores.
    }
    \label{fig:source-results}
\end{figure}

We further compare model performance across DIOR-R, CODrone, and FAIR1Mv2.0, examining fine-grained understanding in general remote sensing, UAV remote sensing, and dense small-object scenarios, respectively.
As shown in Figure~\ref{fig:source-results}(a), most VLMs achieved lower composite scores in dense small-object scenarios than in general remote sensing and UAV remote sensing scenarios.
Figure~\ref{fig:source-results}(b) further shows that all models achieved lower DetA on FAIR1Mv2.0 than on the other two sources. This indicates that instance localization in dense small-object scenarios is a shared challenge across models.
The task-fine-tuned VLMs achieved similar CapA scores across the three scenarios, indicating relatively small differences in their ability to understand and describe instances' own attributes.
For spatial context understanding, RelCap-F1 varied across scenarios differently from DetA. GeoChat, GeoPixel, and EarthDial each achieved higher RelCap-F1 on FAIR1Mv2.0 than in the other two scenarios. They also outperformed the remaining evaluated VLMs on this metric in the same scenario.
By contrast, LocateAnything achieved RelCap-F1 scores of 28.82\%, 23.18\%, and 16.03\% across the three scenarios, respectively. Its localization advantage did not extend to spatial relation understanding, with particularly weak performance in dense small-object scenarios.
These differences must be interpreted alongside instance localization performance. RelCap-F1 is calculated only from spatial relation triples associated with successfully matched instances, whereas all three remote sensing VLMs achieved DetA of at most 1\% on FAIR1Mv2.0. Their higher relation captioning scores therefore reflect caption quality for matched instances and do not imply stronger overall spatial context understanding.

Complex backgrounds, small objects, and dense distributions make object identification and cross-modal alignment more difficult~\cite{kuckreja2024geochat,li2026rsvgzeroov}.
However, image downsampling, visual encoding, and excessive visual token compression may cause the loss of small-object details and local spatial information. Preserving instance details and spatial structure during visual information compression is therefore important for improving fine-grained understanding of remote sensing imagery.
\section{Conclusion}
\label{sec:conclusion}

Referring expression grounding and visual question answering tasks in existing remote sensing vision-language datasets and evaluations typically rely on predefined questions or target referring expressions. They do not adequately assess models' ability to localize instances and describe their attributes and spatial relations without target-specific textual cues.
RICapBench focuses on fine-grained understanding of remote sensing instances and their spatial relations. By combining a spatial relation parser with the composite metric RICScore, it jointly evaluates VLMs' instance localization, attribute captioning, and spatial relation captioning under this setting.
Through a systematic evaluation of 15 advanced VLMs, we examine capability differences and major bottlenecks across these three dimensions. Our findings provide empirical evidence to support improvements in fine-grained understanding of remote sensing imagery and inform future work.

\clearpage
\appendix
\ctexset{section/number=\Alph{section},subsection/number=\thesection.\arabic{subsection}}
\pdfbookmark[0]{Supplementary Material}{supplementary}
\begin{center}
  {\Large\bfseries\papertitle\par}
  \medskip
  {\large\bfseries Supplementary Material\par}
  \medskip
  \begin{tabular}[t]{c}
    \paperauthors
  \end{tabular}
\end{center}
\setcounter{figure}{0}
\setcounter{table}{0}
\renewcommand{\thetable}{S\arabic{table}}
\renewcommand{\theHtable}{supp.\arabic{table}}
\renewcommand{\thefigure}{S\arabic{figure}}
\renewcommand{\theHfigure}{supp.\arabic{figure}}
This supplementary material complements the main paper with further details on RICapBench construction, experimental settings, and model predictions. It is organized into five sections:
\begin{itemize}
    \setlength{\itemsep}{2pt}
    \setlength{\parskip}{0pt}
    \item \textbf{Dataset details.} Section~\ref{sec:appendix-dataset} introduces the three source datasets and examines reference-object expressions in the spatial relation triples extracted from RICapBench captions, reporting their frequency across captions.
    \item \textbf{Prompting strategy.} Section~\ref{sec:appendix-prompting} presents the system, user, and visual prompts used to generate instance captions from source annotations, and explains how the system prompt is adapted to the UAV imagery in CODrone.
    \item \textbf{Spatial relation parser.} Section~\ref{sec:appendix-parser} describes the parser used to extract structured subject--predicate--reference object triples from instance captions for spatial relation analysis.
    \item \textbf{Fine-tuning settings.} Section~\ref{sec:appendix-finetuning} documents the model-specific trainable components, LoRA configurations, and training hyperparameters, together with sequence length, generation limits, and hardware requirements.
    \item \textbf{Qualitative results.} Section~\ref{sec:appendix-qualitative} compares predictions from LocateAnything, GeoPixel, and Qwen3.8-Max with ground-truth annotations across different scenes, illustrating their differences in object localization, instance description, and spatial relation understanding.
\end{itemize}

\section{Dataset Details}
\label{sec:appendix-dataset}

\subsection{Source Datasets}
\label{sec:appendix-source-datasets}
RICapBench draws on three datasets with oriented bounding box annotations, summarized below.

\noindent\textbf{DIOR-R}~\cite{cheng2022aopg}  extends DIOR~\cite{li2020dior} with oriented bounding box annotations for 192,518 instances across 20 object categories. It contains 23,463 images, each measuring $800\times800$ pixels, covering diverse remote sensing scenes and objects such as aircraft, ships, vehicles, and infrastructure.

\noindent\textbf{FAIR1Mv2.0}~\cite{sun2022fair1m} is a fine-grained remote sensing object detection dataset with 37 subcategories organized into five broad categories and annotated with oriented bounding boxes. Its imagery has a spatial resolution of 0.3--0.8\,m. Version 2.0 retains the 16,488 training images of version 1.0, adds a validation set, and expands the test set with data from the 2021 Gaofen Challenge.

\noindent\textbf{CODrone}~\cite{ye2025codrone} contains 10,004 UAV images at $3840\times2160$ pixels, with oriented bounding box annotations for 12 categories. It covers urban, rural, port, and industrial scenes under daytime and nighttime conditions, with imagery captured from multiple flight altitudes and camera angles. The dataset is divided into training, validation, and test subsets in a 50\%/20\%/30\% ratio.

Table~\ref{tab:source-image-counts} reports the final image counts by source and split, including patches retained after cropping and screening.
RICapBench-train uses the source datasets' training splits and the DIOR-R test split. RICapBench-val and RICapBench-evaluation are randomly partitioned from the original validation splits at the source-image level.
The selected images from FAIR1Mv2.0 and CODrone are cropped into $800\times800$-pixel patches and manually screened to remove duplicates, images with ghosting artifacts, and other low-quality samples.
All patches from the same source image remain in a single subset, preventing patches from that image from appearing across training, validation, and evaluation subsets.

\begin{table}[!htb]
    \centering
    \caption{Final image counts by source and split in RICapBench after cropping and screening.}
    \label{tab:source-image-counts}
    \small
    \setlength{\tabcolsep}{3pt}
    \begin{tabular}{@{}lrrrr@{}}
        \toprule
        Source & Train & Val & Evaluation & Total \\
        \midrule
        DIOR-R & 17,600 & 1,999 & 1,164 & 20,763 \\
        FAIR1Mv2.0 & 2,016 & 374 & 274 & 2,664 \\
        CODrone & 2,295 & 640 & 492 & 3,427 \\
        \midrule
        \textbf{Total} & \textbf{21,911} & \textbf{3,013} & \textbf{1,930} & \textbf{26,854} \\
        \bottomrule
    \end{tabular}
\end{table}

\subsection{Category Mapping}
\label{sec:appendix-category-mapping}
To reduce category naming differences across source datasets and avoid matching errors caused by different terms for the same type of remote sensing object, we consolidate the 51 original category labels into 28 semantic categories (Table~\ref{tab:category-mapping}). This mapping groups synonymous or closely related labels and is applied to references to surrounding annotated instances in captions, while preserving the original category annotations.

\begin{table}[!htb]
    \centering
    \caption{Mapping of 51 category labels to 28 semantic categories in RICapBench.}
    \label{tab:category-mapping}
    \small
    \setlength{\tabcolsep}{5pt}
    \renewcommand{\arraystretch}{1.08}
    \begin{tabular}{@{}p{0.25\textwidth}p{\dimexpr0.75\textwidth-10pt\relax}@{}}
        \toprule
        Semantic category & Corresponding category labels \\
        \midrule
        vehicle & vehicle, car, small car, other vehicle, van, bus, truck, cargo truck, dump truck, truck tractor, tractor, trailer \\
        ship & ship, boat, dry cargo ship, liquid cargo ship, engineering ship, other ship, passenger ship, fishing boat, warship, tugboat, motorboat \\
        bridge & bridge, overpass \\
        road junction & intersection, roundabout \\
        baseball field & baseball field \\
        football field & football field \\
        athletics track & ground track field \\
        bicycle & bicycle \\
        motorcycle & motorcycle \\
        tricycle & tricycle \\
        excavator & excavator \\
        airplane & airplane \\
        person & person \\
        airport & airport \\
        harbor & harbor \\
        train station & train station \\
        storage tank & storage tank \\
        chimney & chimney \\
        dam & dam \\
        windmill & windmill \\
        basketball court & basketball court \\
        tennis court & tennis court \\
        golf course & golf course \\
        stadium & stadium \\
        expressway service area & expressway service area \\
        expressway toll station & expressway toll station \\
        traffic sign & traffic sign \\
        traffic light & traffic light \\
        \bottomrule
    \end{tabular}
\end{table}

\subsection{Reference-Object Expressions}
\label{sec:appendix-reference-expressions}

\begin{figure}[!htb]
    \centering
    \includegraphics[width=.75\textwidth]{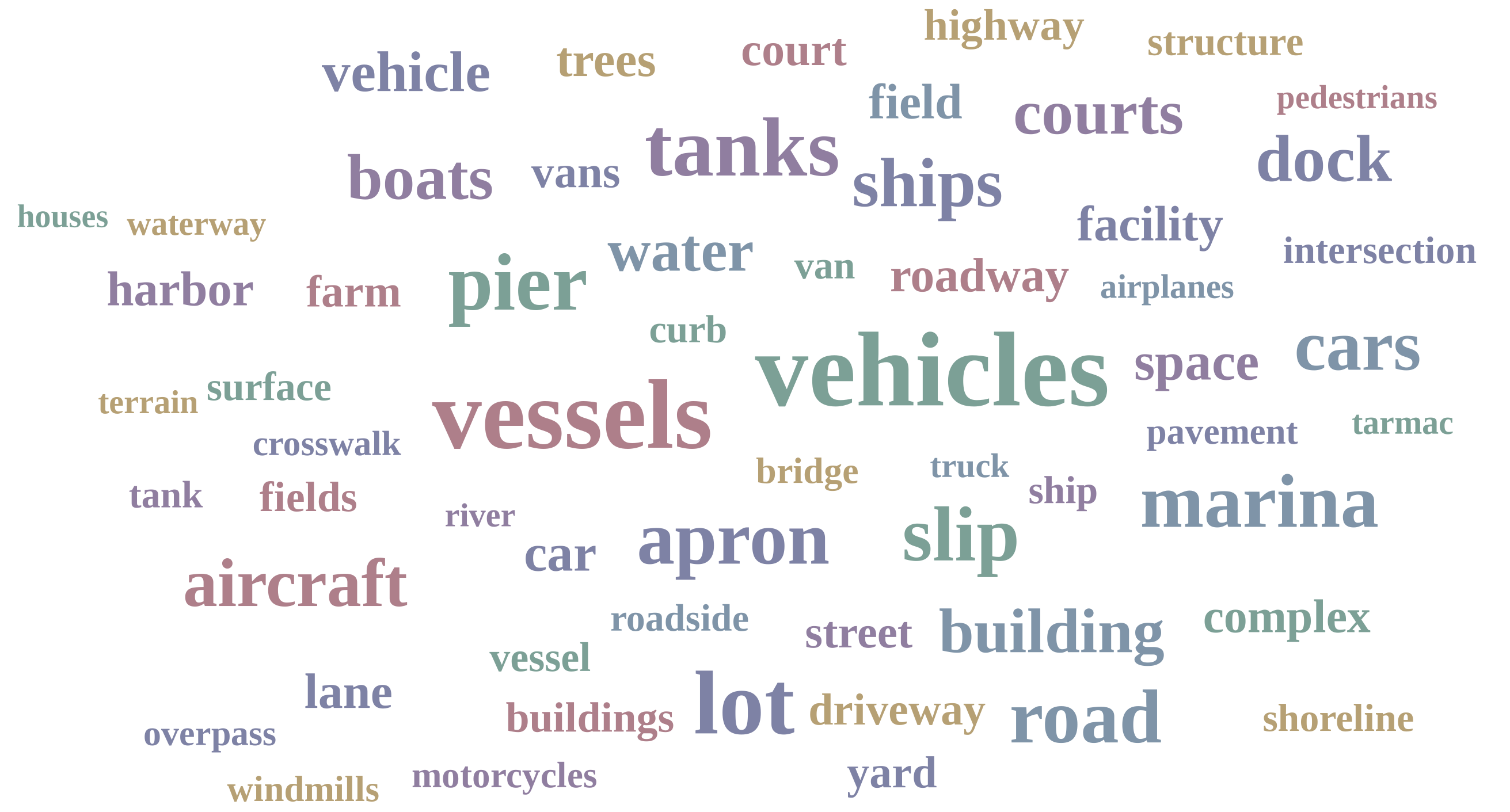}
    \caption{Top 60 reference-object expressions in RICapBench, ranked by the number of captions containing each expression.
    Larger fonts indicate higher caption coverage; colors do not encode semantic categories.}
    \label{fig:spatial-reference-expressions}
\end{figure}

We summarize reference-object expressions in parsed spatial relation triples from all RICapBench captions.
Each expression is counted once per caption, without further merging of singular/plural forms or synonyms.
For example, \emph{vehicle} and \emph{vehicles} are counted separately.
These expressions are distinct from the 28 object categories defined in the main text.

Figure~\ref{fig:spatial-reference-expressions} includes references to discrete instances,
such as vehicles and vessels, and environmental elements, such as roads, piers, and aprons.
This pattern reflects descriptions of both inter-instance relations and relations between target instances and their surroundings.

\section{Prompting Strategy Details}
\label{sec:appendix-prompting}

\begin{figure}[!htb]
    \centering
    \includegraphics[width=.83\textwidth]{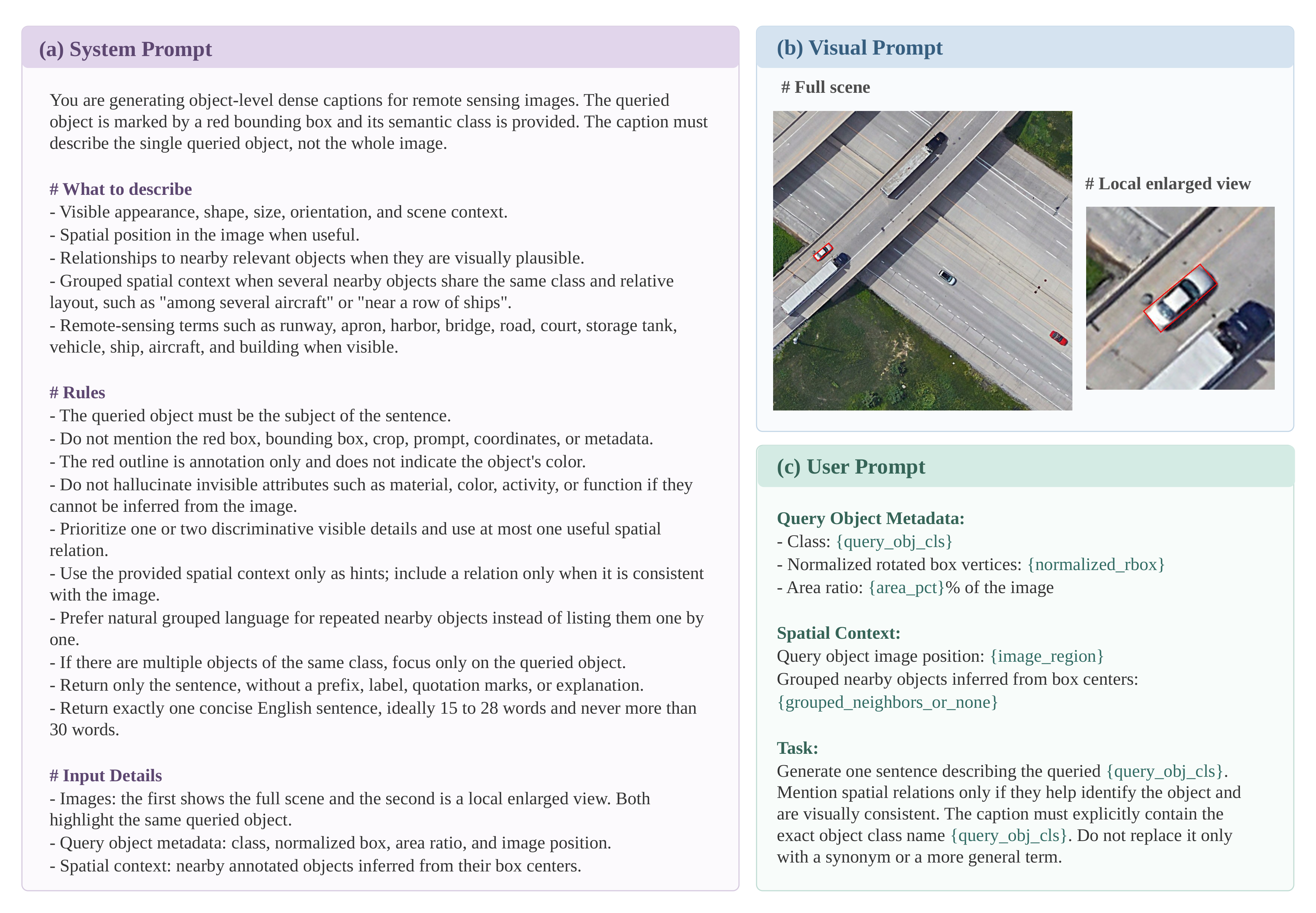}
    \caption{Prompt template for generating descriptions of instance-level spatial relations from source datasets annotated with oriented bounding boxes. The template comprises a system prompt, a user prompt, and a visual prompt.}
    \label{fig:prompting-strategy}
\end{figure}

The prompt template used to construct RICapBench is shown in Figure~\ref{fig:prompting-strategy}.
It comprises a system prompt, a user prompt, and a visual prompt, and uses the oriented bounding box annotations from DIOR-R and FAIR1Mv2.0 to generate instance captions that integrate object appearance and spatial relations. The user prompt specifies the target category, oriented bounding box vertices normalized to a 0–1000 coordinate range, the proportion of image area occupied by the target, and spatial context derived from neighboring objects. Each caption must explicitly include the target category name provided in the prompt, preserving the category terminology of the source dataset, with minor linguistic normalization where needed, to maintain semantic consistency with the source annotations.
The visual prompt provides both the full image and a local crop, with the target highlighted in each, enabling joint interpretation of scene context and object details. The template requires the target instance to be the subject of the caption and includes spatial relations only when they help identify the target and are consistent with the image.

For CODrone, we retain the same user prompt structure and visual prompting procedure while adapting the system prompt to UAV imagery.
 Specifically, we revise the scene vocabulary and relational examples to emphasize roads, intersections, vehicles, pedestrians, and traffic infrastructure, and introduce guidance for varying flight altitudes and overhead and oblique views.
 For small targets, the prompt prioritizes discernible shape, orientation, and the spatial arrangement of neighboring objects. For low-light scenes, it restricts descriptions to clearly visible illumination and colors and prohibits inferring motion, occupants, or object states from a single frame.

\section{Spatial Relation Parser Details}
\label{sec:appendix-parser}

\noindent\textbf{Base rules and triple representation.}
The parser uses Stanford CoreNLP 3.6.0~\cite{manning2014corenlp} for tokenization, sentence splitting, part-of-speech tagging, and lemmatization, followed by explicit extraction rules.
It extracts relations from the first sentence, identifying the target category in the opening subject phrase. Spatial phrases are matched in decreasing order of length and mapped to canonical predicates, preserving expressions such as \emph{to the right of} as \emph{right}.

Each triple contains the target category, a spatial predicate, and a reference object.
Reference objects are represented by physical noun heads from a whitelist covering other instances and environmental features.
Appearance modifiers are omitted, and group expressions such as \emph{a row of trucks} resolve to their member noun, \emph{trucks}. Image-frame references are excluded, and captions with unresolved target subjects yield no triples.

\begin{table}[!htb]
    \caption{Extraction from complex captions beginning with \emph{A vehicle is}. Each output pair gives the predicate and reference object of one triple whose subject is \emph{vehicle}. The symbol \& joins reference members within a single \emph{between} triple; repeated nouns denote separate members.}
    \label{tab:parser-rules}
    \small
    \begin{tabular}{@{}p{.52\linewidth}@{\hspace{.03\linewidth}}p{.45\linewidth}@{}}
        \toprule
        \textbf{Caption continuation} & \textbf{Parser output} \\
        \midrule
        parked with a truck to its left and behind it & (front, truck)\newline (right, truck) \\
        parked with a truck to its left and to its right & No triple \\
        \midrule
        below and to the right of a bridge & (below, bridge)\newline (right, bridge) \\
        between a red and a blue truck & (between, truck \& truck) \\
        beside a road that runs along a river & (beside, road) \\
        \bottomrule
    \end{tabular}
\end{table}

\noindent\textbf{Enhanced parsing rules.}
The two rounds of refinement in Figure~\ref{fig:construction-pipeline}(b) yielded rules for direction, coordination, and clause scope, illustrated in Table~\ref{tab:parser-rules}.
When a \emph{with}-phrase locates a reference object relative to the target, its directions are inverted to retain the target as subject. A diagonal such as \emph{upper left} maps to the single inverse predicate \emph{lower-right}. Within this inversion step, opposing directions for the same reference or alternative directions are discarded.
Active and passive predicates, such as \emph{surround} and \emph{surrounded\_by}, remain distinct to preserve relation direction while keeping the target as subject.

Coordinated predicates sharing a reference noun produce separate triples, whereas \emph{between} combines the reference members into one triple.
Omitted nouns are recovered in expressions such as \emph{a red and a blue truck}. The members are sorted, retaining repeated nouns to represent multiple references of the same type.
Negation and alternative-expression checks filter candidates, while clause markers, verbs, and coordination boundaries restrict extraction to relations attributed to the target. Thus, in the final table example, the vehicle's relation to the road is retained, while the road's relation to the river is excluded.

\smallskip
\noindent\textbf{Relation matching.}
For RelCap-F1, extracted predictions are compared with the ground-truth triples after equivalent facts are deduplicated within each instance.
Matching uses canonical predicates and lexical or synonym-based entity correspondence, allowing endpoint exchange for inverse predicates and symmetric predicates such as \emph{near}.
A maximum one-to-one matching determines $TP_{\mathrm{rel}}$; unmatched predicted and ground-truth triples contribute to $FP_{\mathrm{rel}}$ and $FN_{\mathrm{rel}}$, respectively.

\section{Fine-tuning Settings}
\label{sec:appendix-finetuning}

Table~\ref{tab:finetuning-settings} summarizes the fine-tuning configurations of the models evaluated on RICapBench.
All listed models use LoRA and the AdamW optimizer~\cite{loshchilov2019adamw}, with model-specific trainable components, LoRA parameters, learning rates, and training epochs.
The maximum sequence length is set to 8,192 tokens, and the generation limit is 2,560 new tokens for all listed models.
All experiments involving locally deployed models can be conducted on a system equipped with 2 NVIDIA RTX PRO 6000 GPUs.

For the general-purpose VLMs marked as ALL, we apply LoRA to the vision transformer, MLP projector, and language model to facilitate adaptation to remote sensing.
For the remote sensing VLMs GeoChat, GeoPixel, and EarthDial, we freeze the vision encoder and adapt only the language model.
This choice aims to retain their existing remote sensing visual representations while adapting language generation to the instance captioning task.

\begin{table}[!htb]
    \captionsetup{skip=6pt}
    \caption{Fine-tuning configurations of the evaluated models.
    Trainable components indicate where LoRA is applied. ViT denotes the vision transformer, LM denotes the language model, and ALL includes ViT, the MLP projector, and LM.
    LoRA parameters $r$ and $\alpha$ denote the adapter rank and scaling parameter, respectively.
    Maximum sequence length and maximum new tokens are reported in tokens; the latter limits the number of tokens generated during inference.}
    \label{tab:finetuning-settings}
    \centering
    \small
    \setlength{\tabcolsep}{6pt}
    \renewcommand{\arraystretch}{1.05}
    \begin{tabular}{@{}cccccccc@{}}
        \toprule
        Model & \shortstack{Trainable\\Components} & \shortstack{LoRA~\cite{hu2022lora}\\$(r,\alpha)$} & Optimizer & \shortstack{Learning\\Rate} & Epochs & \shortstack{Max Sequence\\Length} & \shortstack{Output Max\\New Tokens} \\
        \midrule
        Qwen3-VL-2B-Instruct~\cite{bai2025qwen3vl} & ALL & (64,128) & AdamW  & 1e-4  & 10 & 8192 & 2560 \\
        InternVL3.5-2B~\cite{wang2025internvl35} & ALL & (16,64) & AdamW  & 1e-4 & 10 & 8192 & 2560 \\
        Ovis2.5-2B~\cite{lu2025ovis25} & ALL & (64,128) & AdamW  & 1e-4 & 10 & 8192 & 2560 \\
        Gemma-3-4B-it~\cite{gemmateam2025gemma3} & ViT+LM & (64,128) & AdamW & 1e-4 & 3 & 8192 & 2560 \\
        \shortstack[c]{LLaVA-OneVision-1.5\\-4B-Instruct~\cite{an2025llavaonevision15}} & ALL & (64,128) & AdamW  & 1e-4 & 5 & 8192 & 2560 \\
        Qwen3-VL-8B-Instruct~\cite{bai2025qwen3vl} & ALL & (64,128) & AdamW  & 1e-4 & 5 & 8192 & 2560 \\
        InternVL3.5-8B~\cite{wang2025internvl35} & ALL & (16,64) & AdamW  & 1e-4 & 5 & 8192 & 2560 \\
        \shortstack[c]{LLaVA-OneVision-2\\-8B-Instruct~\cite{an2026llavaonevision2}} & ALL & (64,128) & AdamW  & 1e-4 & 5 & 8192 & 2560 \\
        LocateAnything~\cite{wang2026locateanything} & ViT+LM & (64,128) & AdamW & 2e-4 & 3 & 8192 & 2560 \\
        GeoChat~\cite{kuckreja2024geochat} & LM  & (64,128) & AdamW & 1e-4 & 3 & 8192 & 2560 \\
        GeoPixel~\cite{shabbir2025geopixel} & LM & (64,128) & AdamW & 2e-4 & 3 & 8192 & 2560 \\
        EarthDial~\cite{soni2025earthdial} & LM & (64,128) & AdamW & 2e-4 & 3 & 8192 & 2560 \\
        \bottomrule
    \end{tabular}
\end{table}

\section{Qualitative Results}
\label{sec:appendix-qualitative}

\begin{figure}[p]
    \centering
    \includegraphics[width=\textwidth,height=\dimexpr\textheight-36pt\relax,keepaspectratio]{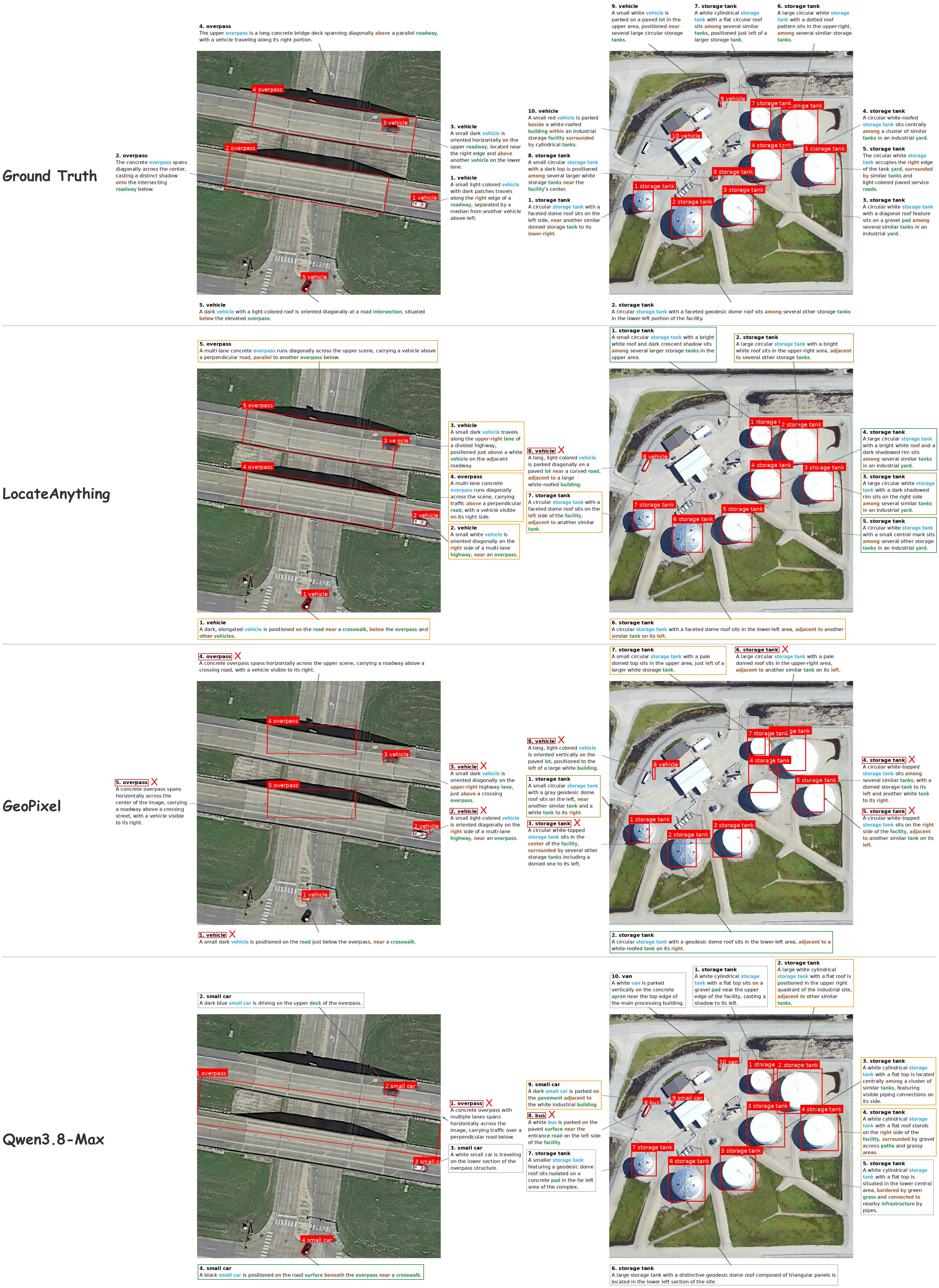}
    \caption{Qualitative comparison of ground-truth annotations and predictions from LocateAnything, GeoPixel, and Qwen3.8-Max on RICapBench images sourced from DIOR-R.}
    \label{fig:qualitative-results-first}
\end{figure}

\begin{figure}[p]
    \centering
    \includegraphics[width=\textwidth]{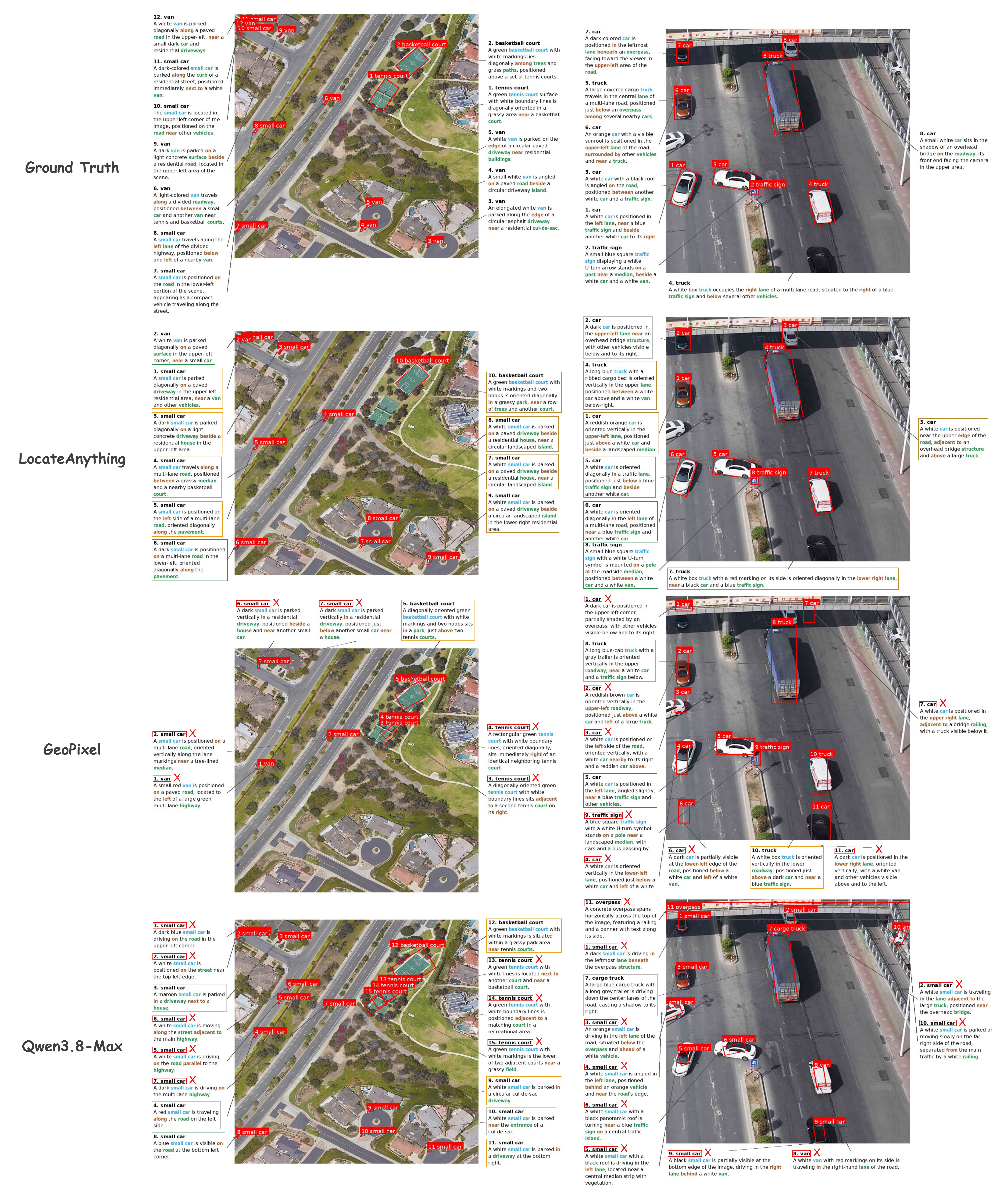}
    \caption{Qualitative comparison of ground-truth annotations and predictions from LocateAnything, GeoPixel, and Qwen3.8-Max on RICapBench images sourced from FAIR1Mv2.0 (left) and CODrone (right).}
    \label{fig:qualitative-results}
\end{figure}

Figures~\ref{fig:qualitative-results-first} and~\ref{fig:qualitative-results} present qualitative results from three representative models, LocateAnything~\cite{wang2026locateanything}, GeoPixel~\cite{shabbir2025geopixel}, and Qwen3.8-Max~\cite{qwen2026qwen38max}, across different scenes in RICapBench.
Each image is accompanied by ground-truth (GT) annotations and instance captions generated by the three models for comparison. We use the spatial relation parser to extract spatial relation triples from each caption and highlight their subjects, spatial predicates, and reference objects in bold sky blue, brown, and green, respectively, illustrating differences in how the models describe target objects and their surrounding spatial context.

Red crosses indicate localization errors. Predicted and GT instances are matched one-to-one, with eligible pairs requiring an oriented bounding box intersection over union (IoU) of at least 0.5.
Unmatched predictions and GT instances contribute to the false-positive and false-negative counts in DetA, respectively. CapA and RelCap-F1 are computed over successfully matched instances.
Green, orange-yellow, and gray caption text boxes qualitatively indicate agreement with the GT captions in overall semantics and spatial relations.
Green indicates strong agreement in both aspects; orange-yellow indicates partial agreement with information omissions or descriptive inaccuracies; and gray indicates poor overall agreement and failure to correctly describe key spatial relations.

In the examples shown, LocateAnything produces bounding boxes that closely align with the GT annotations. Some matched captions omit surrounding context or spatial relations, yielding partial agreement despite successful localization.
GeoPixel's localization errors leave fewer matched instances whose captions contribute to CapA and RelCap-F1. Qwen3.8-Max exhibits both localization errors and incomplete or inaccurate descriptions of object details and spatial relations in these examples.

\clearpage

\clearpage

\end{document}